\documentclass[10pt]{article}  % PSA is 12 point
\usepackage[T1]{fontenc}
\usepackage{tgpagella}
\usepackage{authblk}
\usepackage[top=4cm, bottom=4cm, left=2.75cm, right=2.75cm]{geometry}
\usepackage{footmisc}
    
\usepackage{float}
\usepackage{graphicx}
\usepackage{setspace}
\usepackage{csquotes}
\usepackage{hyperref}
\usepackage{xcolor}
\usepackage{amsfonts}
\usepackage{amsmath}
\usepackage{tikz}
\usepackage[normalem]{ulem}

\usepackage[
    style=authoryear,
    sorting=nyt,
    autocite=inline,
    backend=bibtex
]{biblatex}
\title{Developing a Philosophical Framework for Fair Machine Learning: Lessons From The Case of Algorithmic Collusion}
\author{James Michelson}
\affil{Department of Philosophy, Carnegie Mellon University}

\date{September 25, 2023}
\begin{document}

\maketitle

\begin{abstract}
Fair machine learning research has been primarily concerned with classification tasks that result in discrimination. However, as machine learning algorithms are applied in new contexts the harms and injustices that result are qualitatively different than those presently studied. The existing research paradigm in machine learning which develops metrics and definitions of fairness cannot account for these qualitatively different types of injustice. One example of this is the problem of algorithmic collusion and market fairness. The negative consequences of algorithmic collusion affect all consumers, not only particular members of a protected class. Drawing on this case study, I propose an ethical framework for researchers and practitioners in machine learning seeking to develop and apply fairness metrics that extends to new domains. This contribution ties the development of formal metrics of fairness to specifically scoped normative principles. This enables fairness metrics to reflect different concerns from discrimination. I conclude with the limitations of my proposal and discuss promising avenues for future research.
\end{abstract}

\section{Introduction}

From the perspective of philosophy, the discipline of fair machine learning is a neighbor of other fields in applied ethics like bioethics or business ethics. Unlike these fields, however, the breadth of normative concerns that motivate research is surprisingly narrow. Work outside of the core methodological context of one-shot batch classification problems is sparse \autocite{chouldechova2020}. Yet this is not without justification. Much of the motivation for this methodological focus is driven by the prevalence of issues of discrimination that have resulted from applying machine learning algorithms to social policy problems (for example  \cite{propublica2016, obermeyer2019, chouldechova2017}). One recent survey article \autocite[9]{mehrabi2019survey} even equated fairness with discrimination, defining fairness as ``the absence of any prejudice or favoritism towards an individual or a group based on their intrinsic or acquired traits in the context of decision-making''. Clearly, discrimination is central to the study of fair machine learning and this is reflected in the focus on minimizing disparities between groups. However, the `fair' in fair machine learning has deeper philosophical connections with issues of justice and equality \autocite{binns2018}. Ignoring these issues may turn fair machine learning into an overly narrow field that fails to address the harms arising from new applications of machine learning. Insofar as these new applications lead to injustices beyond discrimination can we bring existing formal definitions and metrics to bear on these problems? If not, how can we develop new formal definitions of injustice and harm in an ethically principled manner? It is these questions this paper attempts to answer.

I explore the case study of `algorithmic collusion' and the associated market fairness issues that it gives rise to. Algorithmic collusion is the idea that algorithms might `learn' to collude, charging prices far above what would be seen in a competitive market \autocite{ezrachi2016}. This does not have a straightforward framing in terms of a canonical fair machine learning classification problem. Furthermore, even if aspects of this problem can be captured by existing work, an entire class of normative concerns---what I call, following \autocite{feinberg1974}, \textit{noncomparative} harms---cannot be captured with existing metrics. The case study of algorithmic collusion explored here is only one example of new fairness problems with entirely different normative concerns than the vast majority of existing problems. Thus, my proposed ethical framework to guide the development of fairness metrics is explicitly designed to capture harms and injustices that are not considered in the existing fair machine learning literature.

My proposed framework is a lexically ordered set of concerns that can be applied to develop fairness metrics that underlie normative principles in any domain. I argue that delimiting the \textit{scope} or \textit{domain} of a fairness problem is the first essential step in this procedure. I believe that domain is best understood substantively instead of methodologically. Problems that involve minimizing disparities in one-shot batch classification tasks across protected groups belong to different domains when they concern different areas of social life. Thus, for example, classification algorithms used to approve loan applications and those used to screen resumes to award interviews do not belong to the same domain: different normative concerns might be operative in each case. Second, once the scope of a fair machine learning problem has been delimited, I believe that practitioners and researchers should specify a \textit{plurality} of harms or injustices that could arise from applying machine learning algorithms in the specified domain. Once this has been accomplished a harm or injustice can then be used as the basis to develop a fairness metric such that its minimization is explicitly tied to mitigating a specific harm.

This paper is organized as follows. In section \ref{case} I introduce the case study of algorithmic collusion and highlight how noncomparative harms that arise in this context fail to be captured by existing fairness metrics. Subsequently, in section \ref{background}, I briefly survey the discipline of fair machine learning, highlighting its philosophically salient features. I then develop my proposed framework in full in section \ref{framework}. In section \ref{defense} I offer a justification for my framework, grounding my arguments in current philosophical debates. In section \ref{application} I apply my framework to developing fairness metrics in the domain of algorithmic collusion. Lastly, in section \ref{discussion} I discuss the limitations of my proposal and give ideas for future research.

\section{Case Study: Algorithmic Collusion}\label{case}

\subsection{Background}

In April 2015 David Topkins became the first e-commerce executive in the United States prosecuted under antitrust law \autocite{newyorker2015}. His crime was to fix the price of posters on Amazon Marketplace in conjunction with a number of other sellers. Topkins and his accomplices all explicitly agreed to use a specific pricing algorithm which would result in charging supra-competitive prices to customers. His case was covered in the popular press with provocative headlines like `When Bots Collude' \autocite{newyorker2015}. This case is notable not for representing a watershed moment in legal precedent but instead for pushing a number of researchers and policy-makers to ask whether algorithms could ‘learn’ to collude without human supervision.

The idea that algorithms might learn to collude, charging prices far above what would be seen in a competitive market, has motivated economists and computer scientists to investigate the phenomenon through extensive simulations (see, for example,  \cite{calvano2020,waltman2008,klein2018,werner2023}). Unlike the case of David Topkins, the future that concerns these scholars is one where ``the industry-wide use of pricing algorithms leads to higher prices, without clear or implied human anticompetitive agreement.'' \autocite[56]{ezrachi2016} In the United States antitrust law does not prohibit firms from jointly raising prices, whether as a response to each other or to an external shock \autocite[9]{harrington2019}. This practice is often harmless. Consider neighboring gas stations located far from competition: if one raises or lowers its prices, the other may soon follow. This practice is often known as `tacit collusion' or `conscious parallelism'. However, what \textit{is} illegal in the United States is coordinated, intentional activity by multiple firms to jointly raise prices. Returning to our gas station example, if these two firms explicitly agreed to raise praises to increase profits then this would be prosecutable under antitrust law. The worry for scholars of algorithmic collusion is that if the phenomenon of supracompetitive prices could arise without either anticompetitive agreement or intent then ``prosecutors now have few, if any, tools to challenge the tacit collusion.'' \autocite[78]{ezrachi2016}

\subsection{Market Fairness and Supracompetitive Prices}

Although not all legal scholars and policy experts are as sanguine about the prospects of algorithmic collusion as the authors cited above \autocite{schwalbe2018, schrepel2017} there is emerging evidence of harm caused by interacting algorithms in electronic marketplaces (see \cite{ezrachi2016} for extensive discussion). For example, in 2011, in what has subsequently become an infamous case of automated pricing gone awry, book sellers on Amazon Marketplace used automated algorithms to set the price of a book in developmental biology, which led to its price reaching more than USD\$20 million. Subsequent analysis suggests this was the result of competing algorithms increasing their price by a fixed multiple based on the price of the competitor algorithm \autocite{eilsen2011}. This comparatively minor case has not to the authors’ knowledge been replicated since. However, an empirical analysis of algorithmic pricing found that about a third of all merchants on Amazon offering approximately 1,600 of Amazon's best-selling products used algorithmic pricing \autocite{chen2016}. This finding---from 2016---is likely an understatement of the current level of algorithmic pricing.

It is no simple matter to repurpose existing fairness metrics for this case of market unfairness. Supracompetitive prices above a competitive baseline affect all market participants, not simply those with certain characteristics. At the level of philosophical principle, we can consider this a \textit{noncomparative harm}\footnote{At risk of confusion with common usage in machine learning, I follow philosophers in using `harm' synonymously with `injustice'.} \autocite{feinberg1974}. Noncomparative harms are those which occur relative to some baseline understanding of human well-being. In contrast, if someone is discriminated against---a clear example of a \textit{comparative} harm---the harm can only be understood ``by reference to [their] relations to other persons'' \autocite[298]{feinberg1974}. Fairness metrics surveyed in \autocite{chouldechova2020,mehrabi2019survey} are constructed to measure disparities between groups and it is far from clear how these might apply to noncomparative harms. Thus, I propose a framework to guide the development of fairness metrics that can equally well be applied to existing problems and entirely new domains like algorithmic collusion. Before developing the framework in full, however, I provide some background on common fairness definitions and highlight their philosophically salient features.

% Market unfairness as supracompetitive prices above a competitive baseline affect all market participants, not simply those with certain characteristics. At the level of philosophical principle, we can consider this a \textit{noncomparative harm}\footnote{At risk of confusion with common usage in machine learning, we follow philosophers in using `harm' synonymously with `injustice'.} \autocite{feinberg1974} which can be understood independently of the harm that befalls others. Comparisons between people are not needed to understand how individuals will be worse off if algorithms collude to raise prices beyond what they can afford\footnote{We discuss later how \textit{comparative} and \textit{noncomparative} harms can coexist in the same problem domain.}. However, it is no simple matter to repurpose existing fairness metrics for this case of market unfairness. Thus, we propose a framework to guide the development of fairness metrics which can equally well be applied to existing problems and entirely new domains like algorithmic collusion. Before developing the framework in full, however, we provide some background on common fairness definitions and highlight their philosophically salient features.

\section{Related Work and Philosophical Background}\label{background}

\subsection{Metrics and Definitions}

Although there are a growing number of definitions of fairness used in the literature I find the high-level distinction between \textit{statistical} and \textit{individual} notions of fairness the most philosophically useful point of departure. The former family of definitions fixes a number of protected groups (i.e., relating to race or sex) and then demands \textit{parity} in some statistical measure across groups. Examples of this work include \autocite{chouldechova2017, calders2020, hardt2016}. This statistical notion is simple in that few, if any, assumptions about the data are made \autocite[85]{chouldechova2020}. However, parity is only aimed for between groups and not individuals. Furthermore, specific statistical group measures of unfairness have been shown to be irreconcilable except under trivial circumstances \autocite{chouldechova2017, kleinberg2017}. Alongside the concern about individual unfairness outlined above, these impossibility results have pushed some machine learning researchers in the direction of individual notions of fairness.

Dwork et al. \autocite*[p7]{dwork2012} document a number of ways in which this simple, statistical notion of fairness may generate statistical parity at the level of groups ``but from the point of view of an individual, the outcome is blatantly unfair.'' Their proposed individual notion of fairness is grounded in the Aristotelian maxim that justice requires `treating similar cases similarly'. In this view, Dwork et al \autocite{dwork2012} ``capture fairness by the principle that any two individuals who are similar with respect to a particular task should be classified similarly''. Unlike the statistical notion of fairness, however, the individual notion of fairness requires significant assumptions. In the case of  \autocite{dwork2012}, the existence of a similarity metric is presupposed which is non-trivial to derive. 

These challenges are serious enough ``that it remains unclear whether individual notions of fairness can be made practical.'' \autocite[85]{chouldechova2020} However, despite practical challenges, this approach attracts ongoing research, perhaps because of its connection to counterfactual fairness. Counterfactual fairness is a fairness metric that is achieved when ``changing [an individual’s assigned treatment] while holding things which are not causally dependent on [the assigned treatment] constant will not change the distribution of the predicted outcome'' \autocite[3]{kusner2017}. These metrics are explicitly causal, which renders them attractive for researchers aspiring to make policy recommendations on the basis of their work.

\subsection{Philosophical Issues in Fairness and Machine Learning}

Much of the work in fairness and machine learning takes place in the long shadow of late twentieth-century Anglo-Saxon political philosophy, most notably under the influence of John Rawls's account of distributive justice \autocite{rawls1971}. In providing a theory of `justice as fairness' Rawls developed an account of justice that gives the concepts of fairness and equality a central place in contemporary political philosophy\footnote{See \autocite{binns2018} for an excellent overview of the core concepts of political philosophy relevant to fair machine learning.}. Leading researchers in machine learning have cited Rawls in defense of their proposed conceptions of fairness (most notably, \cite{dwork2012}), and insofar as the operative understanding of `fair' in fair machine learning is Rawlsian, criticisms of Rawls's work may apply downstream to definitions of fairness favored by machine learning researchers. I focus here on two criticisms of Rawls's theory of justice I believe are relevant to fair machine learning. Firstly, Rawls's focus on principles of justice at the societal level leaves problems for fairness in local contexts. Secondly, Rawls's theory was idealized in such a way as to render reinterpretation of his work for actual use nontrivial and under-specified. 

Rawls \autocite*[p6]{rawls1971} was concerned with developing principles of justice that apply to the ``basic structure of society, or more precisely, the way in which the major social institutions distribute fundamental rights and duties and determine the division of advantages from social cooperation.'' His focus on developing an account of justice for the ``basic structure of society'' led a number of contemporaries to criticize his formulation on the grounds that it was deficient for overlooking unethical individual behavior. More specifically, Rawls's account was charged with overlooking the injustices committed against women in the context of the family \autocite{okin1989} and giving undue weight to individual self-interest in allowing excessive inequalities \autocite{cohen2008}. What is relevant here is that these criticisms all concerned patterns of individual behavior far removed from the context of the ``major social institutions'' of the state and closer to the problem domains of current research in fair machine learning. 

For example, Rawls's first principle of justice encapsulated a ``scheme of equal basic liberties'' \autocite*[p53]{rawls1971} which is designed to apply to fundamental societal issues such as the freedom of speech. It is not clear how to connect this principle to contemporary problems in fair machine learning. Fairness problems in recidivism prediction or predictive policing concern injustices of a much more granular kind than those envisaged by Rawls. These kinds of fairness problems involve specific injustices against specific communities (i.e., higher incarceration rates for African Americans). In appealing to principles of fairness that target the ``basic structure of society'', fair machine learning researchers are failing to ground their ethical worries in principles designed to govern individual actions. It is an open question whether (and when) using fairness metrics that derive normative support from these kinds of principles can be philosophically justified in fair machine learning.

Additionally, Rawls wrote of his account of justice that it was an ``ideal theory'' \autocite*[p8]{rawls1971} and not one suited for regulating our present-day conditions. Insofar as patterns of idealization present in Rawls are found in contemporary research in fairness and machine learning, Fazelpour and Lipton \autocite*{fazelpour2020} have offered three principal criticisms of this `small scale ideal-theorizing'. These are (1) ideal modes of theorizing can lead to systematic neglect of historically situated injustices, (2) ideal evaluative standards do not offer sufficient practical guidance and, (3) these standards fail to delineate responsibilities and liabilities of current decision-makers \autocite[6]{fazelpour2020}. In addition to raising the issue that principles of justice in political philosophy are developed in the context of an ideal theory removed from application, Fazelpour and Lipton prompt important questions concerning the degree of philosophical abstraction desired when conceptualizing fairness in machine learning research.

\section{Proposed Framework}\label{framework}

My proposed ethical framework for developing fairness metrics is a lexically ordered set of concerns, the first of which involves the delimitation of a fairness problem's \textit{scope} or \textit{domain} and the second of which concerns identifying the \textit{plurality} of ethical harms which may arise in the context of a specific scope. By ``lexically ordered'' I mean that concerns of the first sort are to be addressed prior to concerns of the second. I aim to ground disagreements at the level of suitability of fairness metrics in deeper disagreements about the relative merits of different ethical principles (and their associated harms). Using a metric to diagnose or measure unfairness without tying the metric to a harm or principle---however loosely specified---risks removing the metric from consideration as one which can ameliorate the world and rectify some existing unfairness. I defer a defense of my proposal until section \ref{defense}.

I now clarify what is involved in determining the \textit{scope} of a problem of fairness in machine learning. My target is to map fairness concerns onto specific domains of social life so that they can be instantiated by a suitable metric. I consider ``domains of social life'' to be as diverse as college admissions, policing policies, hiring practices, bail policies, market transactions, etc. This non-exhaustive list is intended to capture the idea that in diverse areas of social life, different normative concerns will be operative. Issues of, say, racial discrimination will be more salient in some of these domains, and some normative concerns may be entirely absent in others. I believe similarity at the level of methodology is not enough: similar methodological problems in different domains may require different fairness metrics to capture different normative concerns. I consider two examples. Firstly, a simple case with two very different fairness problems. Secondly, a case where fairness problems that are similar methodologically are dissimilar in their domains of application.

Consider, firstly, two very dissimilar fair machine learning problems. Machine learning algorithms have been used to automate resume screening, determining which candidates should be awarded interviews for a job position based on data from previous successful hires \autocite{dustin2018}. Machine learning algorithms can also be trained on consumers' data to create `differential pricing' \autocite[85]{ezrachi2016} strategies which firms then use to charge different customers different prices to maximize profits. In both examples biases in the training data can lead to injustices when these algorithms are deployed. In the first case, existing hiring practices may reflect the racial or gender biases of interviewers. Although these concerns may be operative in the second case of differential pricing, there are also different normative concerns regarding the privacy of customers' data. It is not that there is no overlap at the level of injustices in these two different cases, only that the sets of concerns are not the same. Therefore, where an injustice in one case will ground the use of a particular fairness definition, that injustice---even if it is operative---may be much less salient in the other case.

Given the prominence of one-shot batch classification tasks in fair machine learning, I also consider the case where this methodology is applied in very different domains of social life. Consider, on the one hand, the problem of predicting recidivism for the assessment of bail \autocite{propublica2016} and on the other, predicting healthcare needs from costs \autocite{obermeyer2019}. In both cases, bias is introduced by differentially good proxy outcomes. In the United States, this may create injustices for African Americans in both cases, although for very different reasons. Both broadly concern issues of racial justice; however, in the former case, there are injustices concerning equitable treatment before the law, and in the latter case concerns of equal access to healthcare are salient. Thus, I believe delimiting the scope of a fairness problem in machine learning is important for isolating relevant harms and injustices. 

Only after the scope of a fairness problem has been identified ought concerns at the level of normative principles be considered. I believe that identifying a number (\textit{plurality}) of harms and injustices is important in recognizing the normative limitations of conclusions drawn from fair machine learning research. For example, returning to the case of differential price discrimination above, I can readily identify issues of (1) discrimination based on gender, (2) discrimination based on race, (3) issues of affordability and inequality, (4) concerns of data privacy, etc. Fairness metrics developed for one of these may be less applicable for measuring or addressing others. Explicitly recognizing this as a normative limitation is helpful for encouraging more accurate policy guidance and recognizing the trade-offs at the level of fairness metrics. I will return to this latter point in section \ref{defense}. I do not believe that a necessarily deep or nuanced understanding of harms or ethical principles is required to address this concern on behalf of researchers of practitioners in fair machine learning. Instead, I believe the purpose of this exercise should be to tie downstream fairness metrics to appropriate normative conceptions of harm and injustice whenever possible.

Once the scope of a problem of fair machine learning has been ascertained and a list of harms identified, the important problems of defining, measuring, and addressing fairness again become central. Crucially, I believe that by adopting my framework, researchers can continue the main lines of their research much as they were before. My framework is designed to limit overgeneralization outside of a specific domain of applicability and sensitize researchers to a plurality of normative concerns operative in their fairness problems.

\section{A Defense of the Framework}\label{defense}

I begin with justifying my focus on the concern of scope. Some notable alternative accounts of justice in political philosophy depart from John Rawls in focusing on local matters of justice at the level of a single institution \autocite{elster1990} or delimited to a `sphere' of human activity, like the family or politics \autocite{walzer1983}. I believe these alternatives to focusing on ``the basic structure of society'' are more relevant to the collection of problems that define the field of fair machine learning. Consider the list of problems that fall under  `local' accounts of justice \autocite[121]{elster1990}: `which workers shall get hired?', `which prisoners shall be released by the parole board?', and `who shall get scarce consumer goods?' My account is inspired by a ``particularist'' \autocite[xiv]{walzer1983} theory of justice which emphasizes that different fundamental principles are operative in different domains of social life.

In contrast, the Rawlsian approach is `universalist' in seeking fundamental principles that do not change depending on the context of application. I believe in the importance of starting from the premise that no single set of principles will apply everywhere in the domain of fair machine learning given its disparate applications. These motivations ground my first \textit{scope} concern. Furthermore, The lexical ordering of the \textit{scope} concern over the \textit{pluralism} concern is a function of the particularist approach just described: it does not make sense to talk about the relevant principles of fairness absent a domain of application. In the case of algorithmic collusion, issues of equitable racial representation---a concern that is operative in the context of my earlier example of automated resume screening---are much less salient.

The motivation for considering a \textit{plurality} of principles of fairness stems from a recognition that different definitions of fairness may conflict with each other. My approach is inspired by the philosophical account of `value pluralism' \autocite{mason2018}. Value pluralism is a philosophical understanding of different values---liberty, fairness, fraternity, etc.---and whether there is some underlying single value they collapse into. Value pluralists believe that different values, or ‘goods’ as is sometimes used in the literature, do not collapse into a single, ultimate value. Like value pluralists, I also believe there is no mechanism---``slide rule'' \autocite[172]{berlin1969}---which can be appealed to adjudicate between competing values. Thus, insofar as, say, statistical and individual definitions of fairness are impossible to reconcile, they can be seen to reflect different ethical concerns at the level of philosophical principle. Furthermore, there is no philosophically justifiable mechanism that can be used to weigh the trade-offs and adjudicate which concern is more profound and has priority. 

Setting aside disagreements about whether a pluralist philosophical position on values is ``the only way to go'' \autocite[4]{cohen2008}, I believe part of the attractiveness of this view is that it offers a lens through which to view impossibility results in fair machine learning \autocite{chouldechova2017, kleinberg2017}. A pluralistic attitude towards values accords with theoretical impossibility results on the irreconcilability of different fairness metrics\footnote{In contrast to Fazelpour and Lipton \autocite[9]{fazelpour2020} I believe that this philosophical account provides a more appropriate interpretation of these theoretical results than merely a ``frank confirmation of the fact that we do not live in an ideal world.''}. Pluralism is motivated by the philosophical recognition that there are many fundamental values on which agreement is hard to reach. On this count, I believe the study of fair machine learning is no different than the rest of applied ethics.

Finally, an upshot of my approach is that it preserves much of the current workflow of contemporary research\footnote{I am unsure to what extent the criticisms of contemporary research practices in fair machine learning by \autocite{fazelpour2020} are mitigated by my approach. Insofar as their target was inappropriate idealization, my `particularist' account is likely to be viewed as a step in the right direction. The relationship is, however, more complicated and beyond the scope of this paper.}. I urge that machine learning researchers and practitioners concerned with fairness better situate their work in philosophical context, explicitly delimiting the scope of their work and the ethical principles (or harms) appealed to. The intended burden on research is in my view fairly minimal and I hope a consequence of adopting this framework is a greater sensitivity to philosophical principles of fairness in machine learning. To demonstrate this, I now turn to the case study of algorithmic collusion to highlight how I envisage the adoption of this framework in an entirely new domain in fair machine learning.

\section{Application of The Framework to Algorithmic Collusion}\label{application}

The setting of algorithmic collusion is far removed from the canonical fair machine learning task of one-shot batch classification. The data are generated from multi-agent reinforcement learning simulations and the phenomenon of interest (i.e., supracompetitive prices resulting from algorithmic collusion) has yet to manifest itself in actual markets so it cannot be measured directly. However, I show that the framework developed in the preceding sections applies equally well to developing metrics that capture our normative convictions in this new setting. I proceed as advocated above: first, I outline the scope of the problem domain. Next, I give an account of ethical or fairness principles that are jeopardized when algorithmic collusion occurs. Lastly, based on the previous two steps, I identify a metric that best captures the resulting unfairness, explicitly tying this metric to both the domain of interest and the ethical principles appealed to.

\subsection{Scope of The Problem}

As outlined in the previous section my description of algorithmic collusion is broad enough to cover a number of scenarios, ranging from algorithmic price wars on a third-party platform like Amazon to the unforeseen consequences of complex high-frequency trading algorithms. Although single companies have already been prosecuted for using algorithms to manipulate markets\footnote{Notably, \textit{Athena Capital Research} paid a USD\$1 million penalty for using its market-manipulation algorithm \textit{Gravy} to overwhelm market liquidity, moving market prices in Athena’s favor (see \cite[68-9]{ezrachi2016} for discussion).}, I have yet to see multiple firms develop algorithms whose combined patterns of behavior result in supracompetitive price manipulations. The lack of cases of algorithmic collusion renders the problem of scope delimitation harder. Indeed, at this juncture, I believe it is still an open question whether the cases that might involve a platform like Amazon marketplace ought to be lumped together with those of high-frequency trading. I remain agnostic on this point and focus on the more studied and discussed case of third-party online marketplaces.

\subsection{Plurality of Harms}

At first glance, it might not be clear what about supracompetitive prices makes it an ethical or fairness issue. We can cast the difference between fairness principles appealed to in the case of discriminative outcomes and those concerning supracompetitive prices as one between \textit{comparative} and \textit{noncomparative} harms \autocite{feinberg1974}. Comparative harms are those that can only be determined in reference to other people, whereas noncomparative harms can be determined independently of others. More concretely, in the case of where African American loan applications are denied where otherwise similar loan applications from white Americans are approved, we can understand this as a comparative harm. The case of supracompetitive prices affects all consumers and may result in net welfare loss \autocite{harrington2019}. This is a noncomparative harm since it may jeopardize the well-being of consumers who depend on competitively priced essential products. Furthermore, insofar as algorithmic collusion results in supracompetitive prices the justification for market mechanisms as a method of allocation and distribution is undermined. Note, the concept of a protected group or class has no traction here and we am forced to develop new concepts and metrics in this domain to make sense of the ethical challenges posed by algorithmic collusion.

There are a number of both comparative and noncomparative harms to consider. It is important to note that the effects of net welfare loss resulting from supracompetitive prices may not be evenly distributed. Wealthier consumers may find their position in the world (relatively) unchanged, whereas poorer consumers might find it much harder to pay high prices for essential items. Thus, it is important to note that comparative and noncomparative harms can inhabit the same domain. Another example of a comparative harm is price discrimination, which might not play out along previously established protected groups (i.e., race or gender) but entirely new ones. These are all equally valid harms worthy of investigation and attention. It is important, however, that I clarify which harm(s) I am concerned with when I move to developing a metric. Thus, I focus on the noncomparative harm of supracompetitive prices and net welfare loss for the remainder of this section.

\subsection{Fairness Metrics}

In order to specify a fairness metric for this phenomenon, a step I recognize as crucial for measuring its occurrence and guiding intervention, I need to first specify a model of the economic market where prices are set. To date, the empirical evidence in support of the plausibility of algorithmic collusion rests entirely on simulations of market agents. Typically, a canonical game-theoretical model of oligopolistic markets \autocite{maskin_tirole_i, maskin_tirole_ii} is used as the setting for multi-agent reinforcement learning simulations. Q-learning algorithms are commonly used (or modified) to represent firms in this market setting. These studies \autocite{calvano2020, klein2018, waltman2008, kimbrough2005, izquierdo2015, dogan2015, werner2023} consider the emergent behavior of reinforcement learning algorithms in these economic games and analyze the degree to which these algorithms learn to charge prices above a competitive baseline.

Economists who have already begun to study algorithmic collusion in these settings have used the metric $\Delta$ of \textit{average profit gain} \autocite[3277]{calvano2020} to measure how far above a competitive baseline prices are set in a given market. The metric is normalized between 0 and 1, where the lowest value reflects a competitive market and 1 reflects a market with a single price-setter (i.e., a monopoly). It is calculated from simulations as

\begin{equation}
    \Delta = \frac{\overline{\pi} - \pi^N}{\pi^M - \pi^N}
\end{equation}

\noindent where $\overline{\pi}$ is the empirical average per-firm profit in simulations, $\pi^N$ is the profit in the Bertrand-Nash (competitive) static equilibrium, and $\pi^M$ is the profit under full collusion (monopoly). This metric can be used to diagnose the extent to which these simulated markets give rise to harms associated with fairness issues in machine learning. I acknowledge this is by no means the only metric that might capture our normative concerns relating to supracompetitive prices, let alone other injustices that may arise in the domain of market transactions. However, this case study represents an end-to-end demonstration of how other such principles can be developed. 

\section{Discussion} \label{discussion}

I have argued that the existing approach to developing fairness metrics in the prevailing methodological context of one-shot batch classification tasks is difficult to extend to new domains. In particular, the normative considerations these metrics capture may not be equally salient in other contexts. My contribution is to develop a framework for guiding fairness research in such a way that aligns the resulting work at the level of fairness definitions and metrics with appropriate normative principles in a philosophically justifiable manner.

In particular, I showed how the framework proposed here applies in the novel context of algorithmic collusion. In this previously unconsidered problem domain, a different set of normative concerns are operative than those in better-studied problems like that of predicting recidivism for bail assessments. I operationalize normative convictions concerning the noncomparative harm of net welfare loss from supracompetitive prices in the fairness metric of average profit gain. This represents a significant step in connecting the field of fair machine learning with the nascent study of algorithmic collusion.

I believe this paper has two principal limitations. Firstly, although I believe any proposal is necessarily incomplete at the level of specifying a plurality of principles, I lack a systematic path to specifying injustices and harms once a domain has been delimited. Developing such an account---if possible---would be invaluable for future research. Secondly, my endorsement of `business-as-usual' research practices in fair machine learning once the framework has been applied risks incurring the criticism of Fazelpour and Liptons' \autocite{fazelpour2020} criticisms of `small-scale ideal theorizing'. 

Future work should address both of these concerns. Additionally, I believe that this framework should be tested in other domains and for other types of machine learning algorithms (i.e., ranking, selection, bandit-learning, etc.) Future work should show where the approach developed here succeeds and fails based on specific features of the new problem domain. My hope is that in doing so, this framework can be amended and improved, ultimately serving to guide ethically motivated research in machine learning and computer science more broadly.

\section*{Acknowledgements}

I would like to thank Sina Fazelpour, David Danks, Nil-Jana Akpinar, Annie Wang, and Giovanna Vitelli.

\section{References} 

\printbibliography[heading=none]

\end{document}